# A²P-MANN: Adaptive Attention Inference Hops Pruned Memory-Augmented Neural Networks

Mohsen Ahmadzadeh[1], Mehdi Kamal[1], Ali Afzali-Kusha[1], and Massoud Pedram[2]
[1]School of Electrical and Computer Engineering, College of Engineering, University of Tehran, Iran
[2]Department of Electrical Engineering, University of Southern California, USA
{mohsen.ahmadzadeh, mehdikamal, afzali}@ut.ac.ir, pedram@usc.edu

*Abstract*—In this work, to limit the number of required attention inference hops in memory-augmented neural networks, we propose an online adaptive approach called A²P-MANN. By exploiting a small neural network classifier, an adequate number of attention inference hops for the input query is determined. The technique results in elimination of a large number of unnecessary computations in extracting the correct answer. In addition, to further lower computations in A²P-MANN, we suggest pruning weights of the final FC (fully-connected) layers. To this end, two pruning approaches, one with negligible accuracy loss and the other with controllable loss on the final accuracy, are developed. The efficacy of the technique is assessed by applying it to two different MANN structures and two question answering (QA) datasets. The analytical assessment reveals, for the two benchmarks, on average, 50% fewer computations compared to the corresponding baseline MANNs at the cost of less than 1% accuracy loss. In addition, when used along with the previously published zero-skipping technique, a computation count reduction of approximately 70% is achieved. Finally, when the proposed approach (without zero-skipping) is implemented on the CPU and GPU platforms, on average, a runtime reduction of 43% is achieved.

## I. INTRODUCTION

Deep Neural Networks (DNNs) are used widely for different applications including image and text classification, as well as speech recognition and other natural language processing (NLP) tasks. As their size (depth and width) grow so as to achieve higher output accuracies, these networks consume more energy due to higher computational complexity along with more memory accesses [1]. A variety of techniques have been proposed to reduce the computational costs, and in turn, the energy consumption of DNNs. Examples of the techniques include weight compression and quantization [2], pruning the weights and connections [3, 4], runtime configurable designs [5, 6], and approximate computing [7].

An emerging type of neural networks is the Memory Augmented Neural Network (MANN), which is based on recurrent neural networks (RNNs). MANNs are highly effective in processing long-term dependent data. Examples of this type of network are those developed by Facebook [8, 9], Neural Turing Machines (NTM) [10], and Differentiable Neural Computers (DNC) [11]. MANNs are equipped with a differentiable associative memory which is used as a scratchpad or working memory to store previous context and input sequences (*e.g.*, sentences of a story) to increase the learning and reasoning ability of the models [12]. This powerful reasoning ability has made the utilization of MANNs common in many application fields including simple dialog systems, document reading, and question answering (QA) tasks [8], [13]–[16]. More specifically, in QA tasks, MANN, first, receives a set of sentences describing a story and the network stores them in its augmented memory. Next, a question is passed to the network which is about the information presented in the story where MANN performs several iterations over an attention-based inference mechanism (called attention inference in the rest of paper) to find the correlation between each story sentence and the question. At the end, this information is employed by a Fully Connected (FC) layer (called the output layer) to generate the answer.

To perform their tasks, different kinds of complex and intensive computations (*e.g.*, dot product calculations) should be performed [17]. The same operations are performed in each layer (*a.k.a.*, hop) of the MANN. Taking multiple hops to greedily attend to different facts is necessary to achieve a high accuracy [8]. The general structure of MANNs for QA tasks is shown in Figure 1.

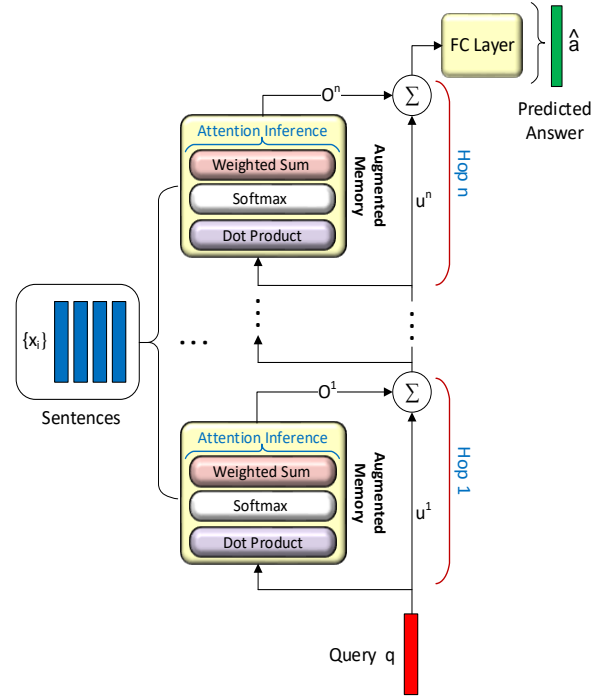

**Figure 1. The general structure of MANNs for QA tasks.**

Embedding the sentences requires $O(n_s d n_w)$ floating point operations (FLOPs) with the same order of memory accesses where $d$ is the embedding dimension while $n_s$ and $n_w$ are the numbers of sentences and the words in each sentence, respectively. Each attention inference hop demands $O(n_s d + d^2)$ FLOPs and the same amount of memory accesses. Also, the final FC layer requires $O(dV)$ FLOPs and memory accesses where $V$ is the size of the dictionary [18]. In addition, the time complexity of the network is $O(n^a)$ ($a \geq 2.375$) implying that the required computations increases superlinearly with the size of the input data (i.e. $n_s$) [17][19]. Therefore, as $n_s$ can increase to thousands/millions of sentences [17] and $V$ can rise to thousands of words [13][18], one should have the computational complexity and number of memory access within reasonable bounds in MANN to make them practical for large real-world problems [18], [20], [21]. Hence, the latency and energy consumption of these networks, especially, in embedded systems, becomes problematic. On the other hand, in the case of IoT applications, the computations may be performed on remote servers. This is feasible only when a communication network is available and also results in low performance [21]. For the above reasons, any effort to reduce the computation burden and energy consumption of MANNs with negligible accuracy reduction is highly desirable.

Since MANN is a recently introduced neural network architecture, the number of works published in the literature focusing, specifically, on reducing the computational complexity of MANNs is limited. To speed up the inference and reduce the operation time of the output layer of MANNs, an inference thresholding method which is a data-based method for maximum inner-product search (MIPS) has been presented in [21]. To reduce the memory bandwidth consumption, reference [17] presented a column-based algorithm with streaming for optimizing the softmax and weighted sum operations, which minimizes the size of data spills and hides most of the off-chip memory accessing overhead. Second, to decrease the high computational overhead, reference [17] introduced a zero-skipping optimization to bypass a large amount of output computation. These works are reviewed in more detail in Section II.

In this work, we present a runtime dynamic technique which adaptively sets the required amount of attention inference hops of MANNs for the input queries in QA applications. The technique employs a small neural network classifier to determine the difficulty level of the input query (*Easy* or *Hard*), based on which the required number of hops to find the final answer is decided. Since the decision is made based on the output of the first hop, a considerable amount of computation may be avoided. To further reduce the energy consumption, we present two approaches for pruning the FC layer of the MANN. The efficacy of the proposed adaptive attention inference hops pruned MANN (called A²P-MANN) is evaluated using two benchmarks including the 20 QA tasks of the Facebook bAbI dataset [22] on End-to-End Memory Networks (MANNs) [8], and 13 question types of the WikiMovies dataset (we will refer to them as *tasks* to be consistent with bAbI tasks) on Key-Value Memory Networks (KV-MANNs) [13].

The remainder of the paper is organized as follows. In Section II, the related work is reviewed. This is followed by a discussion of the structure of MANNs as well as their computational complexity in Section III. We present details of the proposed A²P-MANN inference method in Section IV. Simulation results for the efficacy evaluation of the proposed inference method are given in Section V, and finally, the paper is concluded in Section VI.

## II. RELATED PRIOR WORK

A wide range of approaches to improve the energy and computation efficiency in conventional DNNs (CNNs) have been pursued (see, *e.g.*, [2]–[7]). Interestingly, the intrinsic fault-tolerance feature of NNs allows using approximation methods to optimize energy efficiency with insignificant accuracy degradation [23], [24]. As an example, to reduce the computational complexity, an approximation method based on removing less critical nodes of a given neural network was proposed in [7]. Also, similar pruning techniques have been employed for resource constrained environments like embedded systems [4], [25], [26]. The amount of fault tolerance and resilience varies from one model/structure to another [27]. The efficacy of using approximation on MANNs has been investigated by some prior works (see, *e.g.*, [17], [21]). In [17], three optimization methods to reduce the computational complexity of MANNs were suggested. The first one which was suggested to reduce the required memory bandwidth, was a column-based algorithm that minimized the size of data spills and eliminated most of the off-chip memory access overhead. To bypass a large amount of computations in the weighted sum step during the attention inference hops, zero-skipping optimization was proposed as the next technique. In this technique, by considering a zero-skipping threshold ($\theta_{zs}$), the weighted some operations corresponding to the values less than $\theta_{zs}$ in the probability attention vectors ($p^a$) were omitted. Since the query is only related to a few sentences in the story, a majority of these weighted some operations in the output memory representation step could be skipped using this approach [17]. Finally, an embedding cache was suggested to efficiently cache the embedding matrix.

In [21], a MANN hardware accelerator was implemented as a dataflow architecture (DFA) where fine-grained parallelism was invoked in each layer. Moreover, to minimize the output calculations, an inference thresholding technique along with an index ordering method were proposed. A differentiable memory has soft read/write operations addressing all the memory slots through an attention mechanism. It differs from conventional memories where read/write operations are performed only on specific addresses. Realizing differentiable memory operations have created new challenges in the design of hardware architectures for MANNs. In [28], an in-memory computing primitive as the basic element used to accelerate the differentiable memory operations of MANNs in SRAMs, was proposed. The authors suggested a 9T SRAM macro (obviously different from cell) capable of performing both Hamming similarity and dot products (used in soft read/write and addressing mechanisms in MANNs).

In [20], a memory-centric design that focused on maximizing performance in an extremely low FLOPS/Byte context (called Manna) was suggested. This architecture was designed for DeepMind's Neural Turing Machine (NTM) [10], which is another variant of MANNs, while we have focused on end-to-end memory networks (MemN2N) [8]. Note that these prior works have offered special hardware architectures for MANNs. In this work, however, we propose A²P-MANN technique which is independent from the hardware platform and could be executed on any of these accelerators.

Some runtime configurable designs that give the ability to trade-off accuracy and power during inference have been proposed in the literature [5], [6], [29]. In [5], a Big/Little scheme was proposed for efficient inference. The big DNN (which has more layers) is executed only after the result of the little DNN (with fewer layers) is considered to be not accurate according to a score margin defined for the output softmax. To improve the efficiency of CNNs in computation-constrained environments and time-constrained environments, multi-scale dense networks were suggested in [29]. For the former constraint, multiple classifiers with varying resource demands, which can be used as adaptive early-exits in the networks during test time, are trained. In the case of the latter constraint, the network prediction at any time can be facilitated. A method of conditionally activating the deeper layers of CNNs was suggested in [30]. In this method, an additional linear network of output neurons was cascaded to each convolutional layer. Using an activation module on the output of each linear network determined whether the classification can be terminated at the current stage or not.

### III. MEMORY AUGMENTED NEURAL NETWORKS

#### A. Notation

The following notation is adopted in this paper:
- BOLDFACE UPPER CASE to denote matrices.
- Boldface lower case to denote vectors.
- Non-boldface lower case for scalars.
- The "·" sign is used for the dot product.
- No signs are used for the scalar product or matrix-vector multiplication.

#### B. The Basic Structures of MANNs

MANNs are efficient in solving QA tasks (*e.g.*, bAbI QA tasks) in which the system provides answers to the questions about a series of events (*e.g.*, a story) [8]. A MANN takes a discrete set of inputs $s_1, \ldots, s_n$ which are to be stored in the memory, a query $q$, and outputs an answer $a$. Each of the $s_i$, $q$, and $a$ contains symbols coming from a dictionary with $V$ words. The model writes all $s_i$ to the memory up to a fixed buffer size, and then finds a continuous representation for the $s$ and $q$. The continuous representation is then processed via multiple hops to output $a$ [8]. An example of these tasks is shown in Figure 2. These networks have three main computation phases comprising embedding, attention inference, and output generation.

| s1: | Mary picked up the apple. |
| s2: | John went to the office. |
| s3: | Mary journeyed to the garden. |
| s4: | Mary went to the bedroom. |
| q: | Where was the apple before the bedroom? |
| a: | Garden. |

**Figure 2. An example of story, question, and answer**

In the embedding phase, embedding matrices $A$ and $C$ (which elements are obtained in the training phase) of size $d \times V$ (where $d$ is the embedding dimension and $V$ is the number of words in the dictionary), are used in a bag-of-words (BoW) approach to embed the story sentences into the input and output memory spaces ($M^{IN}$ and $M^{OUT}$ with the size of $d \times n_s$), where $n_s$ denotes the number of sentences in the story. Let $n_w$ denote the number of words per sentence. If the number of story sentences (words of a sentence) is less than $n_s$ ($n_w$), the story (sentence) is enlarged to $n_s$ sentences ($n_w$ words) by zero padding. Thus, for each story, $n_s \times n_w$ words are considered in MANNs.

In the embedding phase, first, each input story is represented by a matrix of size $n_s \times n_w$, which elements are obtained by mapping words of the sentences to their corresponding integer values based on the given dictionary. Next, for each sentence, elements of its corresponding row are used as the row indices of the embedding matrix $A$. The rows of $A$ are chosen to represent each word as a vector of size $1 \times d$. This leads to representing each sentence of size $n_w \times 1$ by a matrix of size $n_w \times d$. To preserve the order of the words in each sentence, and improve the output accuracy, an element-wise product of a positional encoding ($PE$) matrix (of size $n_w \times d$) and the matrix representation of each sentence is performed (details about $PE$ are provided in [8]). Next, elements of the resulting matrix for each sentence are summed along the column direction and transposed, giving rise to the representation of each sentence by a column vector of size $d \times 1$ (called "internal state" of each sentence). These vectors are then stored in $M^{IN}$ to form matrices of size $d \times n_s$ for each story. A similar approach is employed to embed the input story in $M^{OUT}$ by using the embedding matrix $C$. Also, the query (of length $n_w$ words) is embedded by using another embedding matrix $B$ (of size $d \times V$). The output of this phase is the internal state of the query denoted by column vector $u$ (or $u^1$) of size $d \times 1$.

The attention inference phase consists of four parts: inner product, softmax attention, weighted sum, and output-key sum. First, the match between $u$ and each input memory vector (*i.e.*, the internal state of each sentence in $M^{IN}$ which is a column vector of size $d \times 1$) is obtained by computing the vector-matrix dot product

$$k = u^T . M^{IN} \qquad (1)$$

where $u^T$ denotes transpose of $u$ and $k$ is of size $1 \times n_s$. Applying a Softmax operator to the resulting vector gives the probability-attention vector ($p^a$) over the inputs (*i.e.*, $p^a = Softmax(k)$). An output vector $o$ (of size $d \times 1$) is obtained by calculating the sum over the output memory vectors (the internal state of each sentence in $M^{OUT}$ which is a column vector of size $d \times 1$ denoted by $m_i^{out}$) weighted by the corresponding probability value ($p_i^a$ which is a scalar)

$$o = \sum_i p_i^a m_i^{out} \qquad (2)$$

The sum of the output vector $o$ and the query key $u$ ($u^{out} = o + u$) is then presented as the output-key ($u^{out}$) of this phase. In the output generation stage, the network has a fully connected (FC) linear layer in which the final weight matrix $W$ (of size $d \times V$) is multiplied by the output key $u^{out}$ to produce the final answer. Thus, the final output ($\hat{a}$) is obtained from

$$\hat{a} = Softmax(W^T u^{out}). \qquad (3)$$

This means that, to obtain the final output in the training phase, the softmax operation is necessary whereas it may be omitted in the inference phase [21]. This step, independent from the hop counts, is executed once as the last step. To improve the accuracy, MANNs use multiple hop operations [8]. During each of these hops, MANN has a different probability-attention vector over the sentences (facts). Thus, the network concentration is directed towards different facts in each hop that eventually leads to the true output answer [8].

The computations that are performed in each hop depend either on the employed weight tying method or the application type [8], [17], [31]. In addition, different variants of the MANN structure are proposed in literature. Two well-known MANN structures are conventional MANN and Key-Value MANN (KV-MANN). In this work, to realize conventional

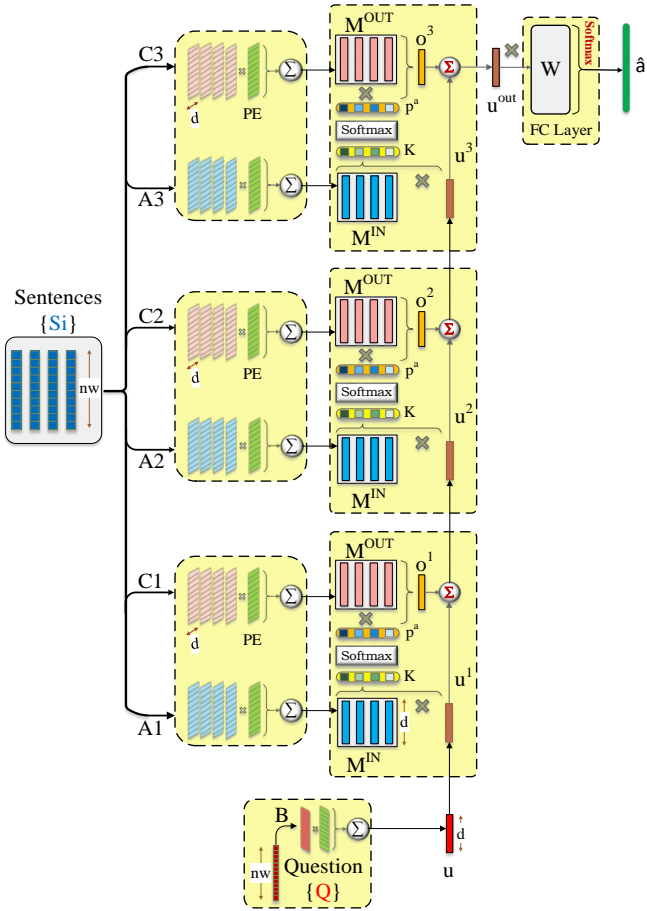
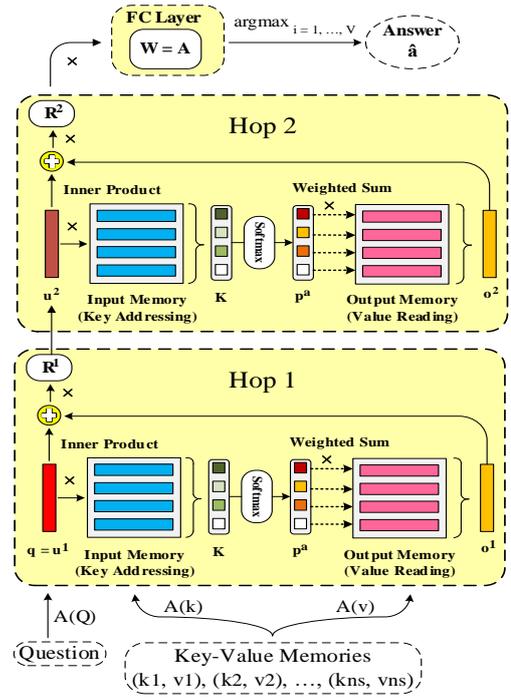

Figure 4. Computational steps of a 2-hop KV-MANN.

Figure 3. Computational steps of a 3-hop conventional MANN.

MANNs, we employ the adjacent weight tying approach [8]. In this approach, the story sentences are embedded in the input/output memory cells using different embedding matrices in each hop (a similar method is used in the Hop-specific approach of reference [32]). This means that each hop contains (assuming the adjacent weight tying approach of [8]) two story embeddings and an attention-based inference. The first hop has also a query embedding. Based on the provided explanations, the general structure of a three-layer conventional MANN, which is used in this work is shown in Figure 3. The structure is based on the MANN architecture described in [8].

In the case of KV-MANNs, some parts of the network structure are different from those of the conventional MANNs. In this case, the input data (*e.g.*, the Wikipedia documents about different movies) is represented in terms of $n_s$ *windows* (instead of $n_s$ sentences) of $n_w$ words. Also, instead of storing sentences in the input and output memories, keys and values are stored in these memories, respectively. In this case, the key includes the features for matching with the input questions while the value consists of the features for matching the responses [13]. For example, the words in the center of the windows or the title of the document (*i.e.*, the name of the movie) may be used as the values and the whole windows are used as the keys. More specifically, in the KV-MANNs, the memory slots are defined as pairs of vectors $(k_1, v_1), \ldots, (k_n, v_n)$. The key windows are embedded to the slots of the input memory and the corresponding value words are embedded to the corresponding slots in the output memory. Thus, the addressing mechanism to the input and output memories in the case of the KV-MANN, is based on keys and values, respectively. The embeddings are simply performed using a BoW approach employing embedding matrices in the KV-MANN (with no *PE* multiplications).

The attention inference phase is mostly the same as that in the conventional MANNs except for the output key generation in each hop. Here, after adding $u$ and $o$, the resulting vector is multiplied by a matrix $R_j$ (which is different for each hop) of size $d \times d$ to generate $u^{out}$ ($u^{out} = R_j \times (o + u)$). Figure 4 shows the structure of the KV-MANN. As shown in the figure, both the key and the value (and also the query sentence) are embedded using the same embedding matrix $A$ [13]. Here, the embedding phase is performed only once in the beginning of the model implementation and the embedded memory slots are used in each hop.

### C. Computational Complexity

The computational complexity of MANNs could be described based on the number of the required floating-point operations (FLOPs). The number of FLOPs for each operation in MANNs is shown in Table 1 where the addition and multiplication operations are counted as 1 FLOP, while each division and exponential are considered as 4 and 8 FLOPs, respectively [33]. During the inference phase in which the input and output memories are usually available, the story sentences are previously embedded to their internal states. Thus, the user only submits question sentences to be answered based on the provided database (*i.e.*, story sentences) [17]. Nevertheless, for interactive applications, the user can provide both the database and the query to be answered [17]. Since, in the latter case, the database is changed for each question, its computational complexity would be significantly larger than that of the former one. We refer to the applications in the former (latter) approach as pre-embedded (interactive) applications. In the case of KV-MANN, since the documents used to answer the questions may be very large (*e.g.*, all Wikipedia pages about movies [13]), we only consider the pre-embedded computations. Now, using the figures and the table, the computational complexity ($CC$) of one hop for pre-embedded ($CC_{H,E}$) and interactive ($CC_{H,I}$) applications may be obtained by

$$CC_{H,E} = \begin{cases} n_s \times (4 \times d + 12) - 1, & \text{conventional MANN} \\ n_s \times (4 \times d + 12) - 1 + 2 \times d^2, & KV-MANN \end{cases} \quad (4)$$

$$CC_{H,I} = n_s \times [(4 \times n_w + 2) \times d + 12] - 1.$$

TABLE 1. NUMBER OF FLOATING-POINT OPERATIONS (FLOPs) FOR EACH OPERATION IN DIFFERENT STRUCTURES OF MANN.

| Operation | FLOPs | Operation | FLOPs |
|---|---|---|---|
| Query/PE embedding | $(2 \times n_w - 1) \times d$ | Softmax attention | $n_s \times 13 - 1$ |
| Query embedding | $(n_w - 1) \times d$ | Weighted Sum | $(2 \times n_s - 1) \times d$ |
| Story/PE embedding | $n_s \times (2 \times n_w - 1) \times d$ | Output Key Sum | $d$ |
| Window/key embedding | $n_s \times (n_w - 1) \times d$ | Output Key Generation | $2 \times d^2 - d$ |
| Story/Query Inner Product | $n_s \times (2 \times d - 1)$ | FC layer | $V \times (2 \times d - 1)$ |

By employing (4), the computational complexity of a conventional three-hop MANN (considering the final FC layer) in the cases of the pre-embedded ($CC_E$) and interactive ($CC_I$) applications are determined from

$$\begin{aligned} CC_E &= (2 \times n_w - 1) \times d + 3 \times CC_{H,E} \\ &\quad + V \times (2 \times d - 1), \\ CC_I &= (2 \times n_w - 1) \times d + 3 \times CC_{H,I} \\ &\quad + V \times (2 \times d - 1). \end{aligned} \quad (5)$$

Also, the computational complexity of a two-hop KV-MANN for pre-embedded applications is obtained from

$$\begin{aligned} CC_E &= (n_w - 1) \times d + 2 \times CC_{H,E} \\ &\quad + V \times (2 \times d - 1). \end{aligned} \quad (6)$$

IV. PROPOSED A²P-MANN

*A. Motivation*

Using multiple computational steps, also called hops or layers, is crucial for improving the average accuracy of MANNs [8]. Very often, two, three or even more hops are utilized. In our study, we observe that a significant portion of the queries can be answered with only one hop not requiring more computation for reaching the correct answer. This motivated us to develop an online mechanism required to identify these inputs (queries), denoted as Easy Input queries, for bypassing the computations of the additional hops at runtime. Obviously, this will result in lower computations and, in turn, less energy consumption per inference. In addition, as mentioned before, the final answer is determined by using a FC layer. Our observation shows that many weights of this layer do not have a great impact on the final result of the MANN. Therefore, pruning this layer leads to a lower latency as well as lower memory usage for storing the weights without sacrificing the output accuracy. Based on these observations, we present the A²P-MANN structure.

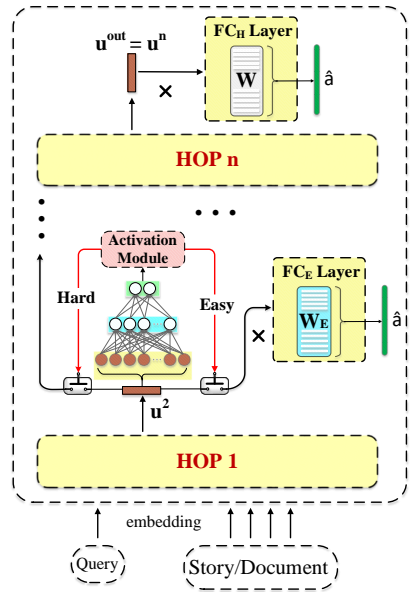

Figure 5. The general structure of the proposed A²P-MANN.

*B. Proposed Structure Overview*

We suggest two methods for the computation reduction of the conventional MANNs. The methods are (i) employing input classifier network (ICN) at the output of the first hop and (ii) pruning the FC layers. In this work, the baseline networks are a jointly trained 3-layer version of conventional MANNs proposed in [8] and a 2-layer KV-MANN proposed in [13]. In A²P-MANN, weights of the baseline MANNs are exploited in shared components. ICN, which is a small network formed by two linear layers, is responsible for classifying the output key of the first hop ($u^2$) to *Easy* and *Hard*. During the inference step, if the ICN identifies the input query as Easy, only one hop is computed, while for Hard-identified input queries, all the hops are applied. For pruning the FC layers, we suggest two approaches to eliminate some rows of the weight matrix of the FC layers (see below). Finally, to further reduce computations in A²P-MANN, one may use the zero-skipping approach suggested in [17]. The general structure of the proposed A²P-MANN is depicted in Figure 5. In the following subsections, the details of the proposed technique are discussed.

*C. Input Classifier Network (ICN)*

When ICN determines that the input query is an easy one, the output of the first hop, $u^2$, is passed to a FC layer (denoted as $FC_E$) trained to classify the easy queries. If ICN finds the input query as a hard one, it passes $u^2$ to the second hop where after computing the remaining hops, the response to the input query is obtained at the output of the FC layer (denoted as $FC_H$). The FC layer uses the output of the final hop. In the proposed approach, $FC_H$ is the same FC layer as that of the baseline MANNs.

**$FC_E$ layer:** The architecture of $FC_E$ is the same as the $FC_H$, while the weights of the $FC_E$ ($W_E$ as the corresponding weight matrix) are trained based on the output of the first hop. In this work, we considered the weights of the $FC_H$ as the initial weights for the training of $FC_E$. In the training phase, $FC_E$ is trained while the pretrained embedding matrices are frozen. This layer is trained by the all the input queries from the training set of the considered dataset. Also, the FC$_E$ layer is trained before training the ICN.

**ICN Labels:** Because the ICN's goal is to classify the queries into easy and hard, for each query of the considered dataset, it should generate a label. For this, each story-query input is fed into an evaluation process using both 1-hop and multiple-hop (3 for the conventional MANN and 2 for the KV-MANN) states of the network. In this process, when the FC$_E$ layer provides correct response, label *Easy* is considered for that input query. If the FC$_E$ output is not correct and FC$_H$ provides the correct answer, *Hard* is assigned as the label for the input query. Note that if neither of the states could predict the true label for an input, we assign label *Easy* to gain more computation reduction without sacrificing the output accuracy.

**ICN Architecture:** This network should be designed based on the considered dataset and the parameters of the baseline MANN. In this work, we designed two ICN networks which are mostly similar; one based on 20 tasks of bAbI, which is a commonly used dataset for the conventional MANN (see, *e.g.*, [8], [17], [20], [21], [28], [32]), and the other is based on the WikiMovies dataset which is mostly used for evaluating KV-MANNs [13]. Our investigation shows that implementing ICN using small linear layers can provide an acceptable accuracy. In this work, a two-linear-layer network achieves up to 86% (84%) accuracy in classifying input queries for MANN/bAbI (KV-MANN/WikiMovies) benchmark. The first layer of this network transfers the input ($u^2$) of size *d*, which, in our study, is 40 for the conventional MANN and 500 for the KV-MANN, to a hidden layer of fewer neurons (32 for the conventional MANN and 200 for the KV-MANN). Here, ReLU is used as the activation function of the hidden layer neurons. The hidden neuron outputs are then transferred to the output layer of 2 neurons for labeling the tasks as *Easy* or *Hard*.

**Using an Activation Module:** Because the accuracy of ICN may not reach 100%, having false-positive inputs (*Hard* inputs that are identified as *Easy* by ICN) is inevitable. These incorrect classifications will lead to A²P-MANN accuracy degradation. Using the activation module of [30] at the output layer of the ICN can be helpful. The said module utilizes the confidence value (probability value, *z*) at the output neuron as well as a predefined confidence value threshold to decide whether the input query should be classified as *Easy* or *Hard*. More specifically, the activation module is an "if statement" which compares the maximum of the two outputs of ICN (logits/probabilities) with a threshold. When the maximum value (*i.e.*, the confidence value) belongs to the *Easy* output and is larger than the threshold, the implication is that the input is easy and invokes FC$_E$ to compute the answer after the first hop. If the confidence value is less than the threshold, the answer is determined by performing the computations for all the hops. Utilizing this module has the drawback of increasing the number of false-negative inputs (*Easy* inputs that are identified as *Hard* by ICN) and consequently, abandoning some of the computational savings. Fortunately, this increase does not deteriorate the accuracy. Based on the above explanation, the output of ICN should pass through a softmax layer before entering the activation module.

### D. Pruning Weight Matrices of FC Layers

Owing to the large dimension of MANN output (which is the number of words in the dictionary), fine-grained parallelism is a challenge for the computations in the last FC layer. Methods such as serializing some dot products by defining an inference thresholding method and therefore ignoring some dot products of this layer have been suggested in literature (i.e., [21]).

Normally, it is not possible to find a QA dataset/task in which all the words (in the dictionary) are among the labels (answers). That is because, for example, many of the words in the dictionary, such as prepositions, question words, articles, and some other determiners are stopwords. Our investigation shows that for the bAbI (WikiMovies) dataset, more than 65% (85%) of the words in the dictionary are not among the training labels, and thus, are never the answer to any question in the training data. In other words, the MANN architecture is not trained to pick up the answer among them. This is a direct consequence of the fact that a limited number of words from the dictionary are used as the answers for the questions. Hence, we suggest that the corresponding rows for those words (called *unused* rows) in the final weight matrices ($W$ and $W_E$) be eliminated in the inference phase. This should not have any adverse effects on the final accuracy. In addition, we suggest to prune *unimportant* rows from the final weight matrices. This pruning leads to a very small accuracy loss. Setting a pruning threshold, symbolized by $\theta_P$, unimportant rows are defined as the ones in which a notable number of weights, denoted by $N_P$, have absolute values less than $\theta_P$. By applying these two pruning approaches, a significant portion of the output calculation is eliminated and application of fine-grained parallelism to the operations of the output layer may become possible. In some cases, the two proposed pruning approaches may suggest pruning of the same row(s), depending on the considered values of $\theta_P$ and $N_P$, which are functions of the maximum tolerable accuracy loss. It should be noted that considering larger (smaller) values for $N_P$ ($\theta_P$) would result in less accuracy loss. Note that, in this work, we use the same values of $\theta_P$ and $N_P$ for pruning in both cases of FC$_H$ and FC$_E$.

To perform the pruning, first, we find the unused or unimportant indices which is used to create a list for the important indices (all the indices not among the unused or unimportant indices). Based on this, we define a new matrix which is only made of the rows corresponding to the important indices. During the inference, instead of the original redundant matrix, we use this pruned matrix for the final FC layer. The predicted index at the output is then given to the list of important indices to choose the final predicted index (answer).

It is worth mentioning that while *unused* and *unimportant* rows are useless in the FC layers, they should be considered in the attention inference hops as explained next. During the output generation step (FC layers), the network finds some type of similarity between each row (corresponding to a word) and the output key $u^{out}$. Thus, in this way, the word with the maximum similarity will be the maximum logit or the answer. The corresponding word indices of the rows that we prune in the final weight matrices of the FC layer are not among the training labels. Therefore, during the training phase, the weights in those rows in $W$ (and $W_E$) have small absolute values leading to the least similarity with the output key of the final hop. Hence, they yield the lowest logits making them unimportant in the FC layer. These rows, however, are important for the embedding and attention-based inference step, and hence pruning them in the embedding matrices will lead to some accuracy degradation. This is because when producing the attention vector, the network finds the degree of similarity between each sentence of the story and the question making every word important in those steps.

### E. Computation Reduction of $A^2P$-MANN

Based on the provided details of the proposed $A^2P$-MANN, the reduction in the computational complexity (CR or Computation Reduction in FLOPs) of the models for the two cases of the pre-embedded (in both the conventional MANN and KV-MANN structures) and interactive (only for the conventional MANN) applications could be obtained, respectively, using

$$CR_E = \zeta_E \times (m-1) \times CC_{H,E} + P_R \times [V \times (2 \times d - 1)] - [2 \times L_1 \times (d+2) + 25]$$

$$CR_I = \zeta_E \times 2 \times CC_{H,I} + P_R \times [V \times (2 \times d - 1)] - [2 \times L_1 \times (d+2) + 25]$$
(7)

where the parameter $\zeta_E$ is the probability of an input query to be classified as an *Easy* input by ICN. Also, $m$ is the number of hops (here, 3 for the conventional MANN and 2 for the KV-MANN). Recall that $V$ denotes the number of words in the dictionary. $P_R$ is the ratio of the pruned rows to total number of rows in the final FC layers (*i.e.*, FC$_H$ and FC$_E$). The last terms of the provided formulas show the overhead of the considered ICN network, where $L_1$ is the number of the ICN hidden layer neurons. Obviously, when an ICN with more hidden layers is employed, the considered ICN overhead has to be modified.

By applying the zero-skipping approach to $A^2P$-MANN, more computation will be eliminated. In this case, the computation reduction of the $A^2P$-MANN may be determined as follows

$$CR_E = \zeta_E \times (m-1) \times CC_{H,E} + P_R \times [V \times (2 \times d - 1)]$$
$$+ ((\zeta_E \times \Psi_E + (1-\zeta_E) \times m \times \Psi_H) \times [(2 \times n_s - 1) \times d])$$
$$- [2 \times L_1 \times (d+2) + 25]$$

$$CR_I = \zeta_E \times 2 \times CC_{H,I} + P_R \times [V \times (2 \times d - 1)]$$
$$+ ((\zeta_E \times \Psi_E + (1-\zeta_E) \times 3 \times \Psi_H) \times [d \times (n_s \times (2 \times n_w + 1) - 1)])$$
$$- [2 \times L_1 \times (d+2) + 25]$$
(8)

where, $\Psi_E$ ($\Psi_H$) is the average ratio of the number of skipped values to $n_s$ (the total number of values in the probability-attention vector) for *Easy* (*Hard*) input queries in the $A^2P$-MANN.

## V. RESULTS AND DISCUSSION

### A. Simulation Setup

The efficacy of the $A^2P$-MANN inference technique was assessed on the conventional MANN using 20 QA tasks of the bAbI question and answer dataset [22] as well as on the KV-MANN using different question types (tasks) of the WikiMovies dataset [13]. The bAbI testset consisted of 20,000 samples where each involved a story ($n_s$ sentences), a

**TABLE 2. THE CONSIDERED PER-TASK CONFIDENCE VALUE THRESHOLDS ($z_{pt}$) FOR THE ICN TRAINED FOR BABI TASKS.**

| Task | P | Task | P | Task | P | Task | P |
|---|---|---|---|---|---|---|---|
| 2* | 0.52 | 8 | 0.68 | 12 | 0.8 | 18 | 0.99 |
| 3 | 0.8 | 9 | 0.85 | 14 | 0.84 | 19 | 0.59 |
| 6 | 0.95 | 10 | 0.99 | 15 | 0.67 | | |

* The tasks that are not mentioned did not need an activation module in the ICN.

**TABLE 3. THE CONSIDERED PER-TASK CONFIDENCE VALUE THRESHOLDS ($z_{pt}$) FOR THE ICN TRAINED FOR WIKIMOVIES QUESTION TYPES.**

| Task(s) | P |
|---|---|
| movie to genre – movie to tags – movie to writer | 0.85 |
| movie to actors | 0.92 |
| actor to movie – director to movie – movie to director | 0.95 |
| movie to language – writer to movie | 0.98 |

* The tasks that are not mentioned did not need an activation module in the ICN.

query, and a label (answer) as shown in Figure 2. Each task had 1,000 samples implying 20 tasks for the samples. As stated previously, the considered conventional (KV-) MANN was a jointly trained 3(2)-layer version of the (KV-) MANN with adjacent weight tying and positional encoding (PE) multiplications [8]. The parameters $d$ (embedding size), $n_s$ (number of sentences), and $V$ (vocabulary size in dictionary) was, 40, 50 and 174, respectively. The WikiMovies testset included ~10K sample questions in the movie domain which can be broken down into 13 classes of different question types. In the KV-MANN, before inferencing, the facts (in this work the Wikipedia documents about movies) are stored in a key-value structured memory. In our experiments, the parameters $n_s$ (the memory size or the number of the windows for the documents), $d$, and $V$ were 50k, 500, and ~238k, respectively.

Also, as mentioned before, the proposed ICNs for the considered networks were consisted of two linear layers in which the first layer (hidden layer) had 32 (200) neurons for the MANN/bAbI (KV-MANN/WikiMovies) and the second layer had 2 neurons. For both considered MANNs, the ICNs were trained jointly on all the tasks and a weighted cross entropy loss function was used. In addition, the Adam optimizer with a learning rate of 0.01 (0.001) was invoked for training the ICN in case of the MANN/bAbI (KV-MANN/WikiMovies). Since the quality of answers in the MANN structures strongly depends on hyperparameters tuning in its training process, the training quality of ICN is dependent on the (pre)trained embedding matrices [8]. This had us to utilize the joint version of the TensorFlow source code of [34] for the conventional MANN in order to determine the pretrained matrices. Also, the Torch implementation of the KV-MANN [35] was reimplemented in Tensorflow for determining the pretrained embedding matrix. Our proposed $A^2P$-MANN, however, was implemented in PyTorch [36] for the evaluation. In this work, the bAbI dataset with 1k training problems per task for training the considered ICN was used. In the case of the KV-MANN/WikiMovies, since the number of samples was different in each task, different numbers of samples from each task were included to train the ICN.

To improve the ICN accuracy, confidence-value thresholds under two cases of a global threshold for all the tasks (*i.e.*, $z_G = 0.6$ for the MANN/bAbI and $z_G = 0.85$ for the KV-MANN/WikiMovies) and per-task thresholds (*i.e.*, $z_{pt,i}$ where $i$ refers to the $i^{th}$ task) were considered. Table 2 reports the $z_{pt}$ for each task of the bAbI dataset. For some tasks (*e.g.*, the first task), since the activation module did not improve the accuracy of ICN, it was omitted from ICN, and hence, confidence thresholds have not been reported for them. Also, we have reported these values for WikiMovies dataset in Table 3. These thresholds were found empirically using the training datasets and depended on both pre-trained embedding matrices and ICN trained layers.

We have suggested two pruning approaches for the final FC layers including eliminating the *unused* and *unimportant* rows. For the latter, in the case of MANN/bAbI (KV-MANN/WikiMovies) benchmark, the values of 0.1 (0.05) and 13 (240) for the parameters $\theta_P$ and $N_P$, respectively, were considered which led to no accuracy degradation. Also, more than 90% of the selected *unimportant* rows were the *unused* rows. Finally, by increasing $\theta_P$ (and/or decreasing $N_P$) more rows are omitted, degrading the accuracy.

## B. Results and Discussion

***ICN efficacy:*** The percentages of input queries, in each task, detected as *Easy* by ICN for three confidence threshold scenarios are illustrated in Figure 6. The scenarios included considering the confidence values of $z_{pt} = 0.5$ (NC), $z_{G-bAbI} = 0.6$ or $z_{G-WikiMovies} = 0.85$ (Global), and a separate $z_{pt}$ for each task (PerTask). The results, on the bAbI tasks, indicate that, in the 1[st] and 20[th] tasks about 100% of the input queries were classified as *Easy* in all the scenarios, while in the worst case, less than 2% of the input queries of the 2[nd] task were determined as *Easy* input queries in different scenarios. The last three bars in this figure show the average percentages of the *Easy* input queries among all the tasks (denoted as *Avg*) in all the three considered scenarios. It reveals that 75% of the inputs were considered as *Easy* ones when the activation module was not employed (the NC scenario), while 67% (63%) of the inputs were detected as *Easy* in the Global (PerTask) scenario. On the WikiMovies dataset, nearly all of the inputs in the "movie to imdbrating", "movie to imdbvotes" and "movie to year" tasks were *Easy* in some scenarios while less than 50% (and even less than 30%) of the inputs were classified as *Easy* for "movie to actor" and "tag to movie" tasks in different scenarios. On average, 74% of the inputs were considered as *Easy* ones for the NC scenario, while 63% (59%) of them were classified as *Easy* for the Global (PerTask) scenario. In the case of the KV-MANN/WikiMovies, the reported average values were calculated by weighted averaging of the data due to the different numbers of samples in the tasks.

Also, the ratio of the inputs that classified as *Easy* decreased when the activation module was employed. This originated from the fact that the false positive (negative) classifications were reduced (increased) in the Global and PerTask scenarios. Figure 7 depicts the accuracy of ICN for the three scenarios based on considered confidence threshold values. The results reveal that considering these confidence threshold values leads to the overall accuracy reduction of ICN (due to more false negative inputs). Also, it is evident that in the case of the PerTask scenario, the accuracy decrease for ICN is more than the Global scenario. This is due to the use of the activation module for reducing the false positive (FP) metric of ICN which leads to the accuracy improvement of A[2]P-MANN. The results suggest that the FP metric of ICN in PerTask scenario be the lowest among the three scenarios. Table 4 and Table 5 contain the minimum, maximum, and average false positive and false negative (FN) metrics of the ICN network for different scenarios of the MANN/bAbI and KV-MANN/WikiMovies benchmarks. The results show that the FN of the proposed ICN is larger than its FP (on average, by ~4×). Apparently, FN does not impact the final accuracy of the A[2]P-MANN. The FP of the ICN when the activation module is employed was reduced. As was expected, the lowest FP belonged to the PerTask scenario. The accuracy of the first (second) ICN (without considering the confidence threshold) was about 100% (98%) in the case of the 1[st] and 20[th] tasks ("movie to year" question type) of the bAbI (WikiMovies) dataset while the lowest one belonged to the 3[rd] ("tag to movie") task with the accuracy of 66% (56%). On average, the accuracy of the first (second) ICN was 86% (84%), 81% (78%), 79% (76%) for the MANN/bAbI (KV-MANN/WikiMovies) benchmark in the cases of NC, global, and PerTask scenarios, respectively.

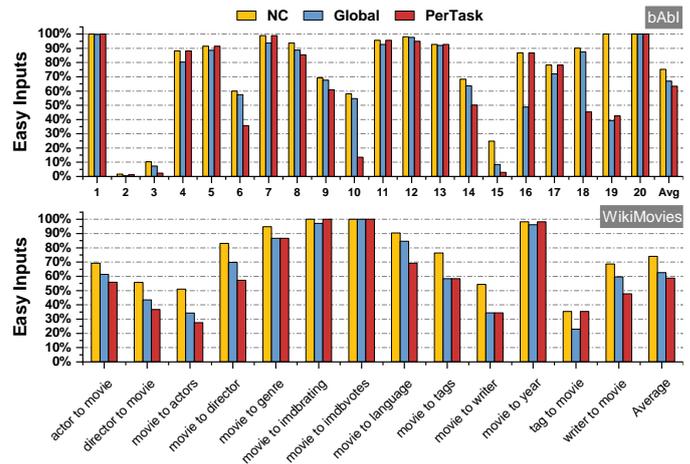

Figure 6. Percentage of Easy-classified inputs using ICN under three considered scenarios for the confidence value.

Finally, the A[2]P-MANN accuracy is compared to that of the conventional MANN and KV-MANN structures under the Global and PerTask scenarios. The results of this comparison are plotted in Figure 8. As shown for the bAbI tasks, in the worst case, a 4.6% accuracy loss in task 14 was yielded, while in some cases (*e.g.*, tasks 11 and 13) the accuracy was improved up to 2.7%. On the other hand, for the WikiMovies tasks, the worst case corresponded to a 3.4% accuracy loss in the "director to movie" task while in some cases (*e.g.*, "movie to imdbrating" and "movie to imdbvotes") up to 1.1% accuracy improvement was obtained. As stated before, the considered parameters for the pruning techniques (as well as Zero-Skipping) had no adverse effect on the accuracy. This implies that the degradation should be attributed to the input classifier network and $FC_E$ efficiency. In few tasks, the proposed approach leads to a higher accuracy compared to that of the conventional (KV-)MANN. The improvement is attributed to the retrained FC layer used after the first hop in the proposed approach. Applying this retrained layer on the output of the first hop is a different structure than that of the original structure leading to a correct answering of some queries to which the conventional (KV-)MANN provides wrong answer.

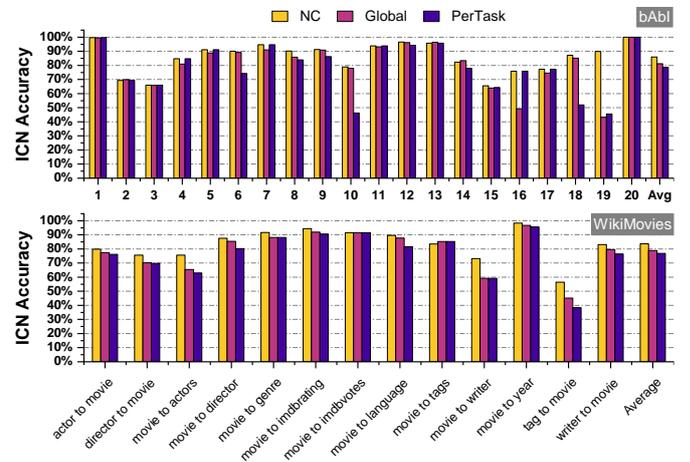

Figure 7. Accuracy of ICN for the considered scenarios.

**TABLE 4. THE MINIMUM AND MAXIMUM FALSE POSITIVE AND FALSE NEGATIVE METRICS OF ICN IN bAbI TASKS FOR THREE CONSIDERED SCENARIOS.**

| | Considered Scenario | Minimum (corresponding task(s)) | Maximum (corresponding task(s)) | Average |
|---|---|---|---|---|
| False Positive | NC | 0% (20) | 13.5% (16) | 4.78% |
| | Global | 0% (20) | 8.7% (17) | 3.73% |
| | PerTask | 0% (20) | 13.5% (16) | 3.19% |
| False Negative | NC | 0% (1,19,20) | 42.8% (3) | 9.21% |
| | Global | 0% (20) | 53.7% (19) | 15.27% |
| | PerTask | 0% (1,20) | 50.9% (19) | 17.81% |

**TABLE 5. THE MINIMUM AND MAXIMUM FALSE POSITIVE AND FALSE NEGATIVE METRICS OF ICN IN WIKIMOVIES TASKS FOR THREE CONSIDERED SCENARIOS.**

| | Considered Scenario | Minimum (corresponding task) | Maximum (corresponding task) | Average |
|---|---|---|---|---|
| False Positive | NC | 0.6% movie to year | 9.3% (director to movie) | 5.1% |
| | Global | 0.5% movie to year | 8.3% (movie to imdbvotes) | 3.5% |
| | PerTask | 0.6% movie to year | 8.3% (movie to imdbvotes) | 3.0% |
| False Negative | NC | 0.1% (movie to imdbrating) | 45% (tag to movie) | 11.2% |
| | Global | 0.1% (movie to imdbvotes) | 56.2% (tag to movie) | 17.7% |
| | PerTask | 0.1% (movie to imdbvotes) | 59.6% (tag to movie) | 20.1% |

It should be noted that considering different confidence values for the tasks (the PerTask scenario) results in a higher accuracy for $A^2$P-MANN compared to the Global scenario. In the PerTask scenario, we tried to limit the accuracy loss to near 1% for each task. On average, the PerTask (Global) scenario led to 0.1% (1%) $A^2$P-MANN accuracy loss compared to that of the baseline MANN on bAbI tasks. Also, on average, among WikiMovies tasks, $A^2$P-MANN had an accuracy loss of 0.9% (1.5%) compared to that of the baseline KV-MANN in the PerTask (Global) scenario.

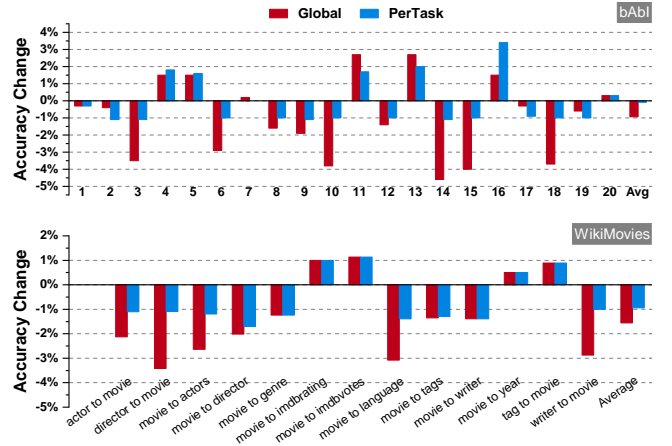

**Figure 8.** Accuracy loss of $A^2$P-MANN compared to the conventional MANN (upper chart) and KV-MANN (lower chart) structures under two considered scenarios.

*Platform Independent Performance Evaluation:* To determine the performance improvement of $A^2$P-MANN (with and without using the zero-skipping technique [17]) compared to the conventional MANN and KV-MANN for Pre-Embedded and Interactive application types, we have utilized (5)-(8). Figure 9 and Figure 10 show the FLOPs reductions (performance improvement) of $A^2$P-MANN in different tasks of the bAbI and WikiMovies datasets, respectively, in the Global and PerTask scenarios. Also, in this figure, the reductions due to using the zero-skipping technique for the baseline MANNs are provided. The parameter $\theta_{ZS}$ was considered as 0.01, providing an 81% overall reduction in the weighted sum operations without any accuracy loss [17]. More precisely, our results on bAbI indicated that 40% (23%) of the achieved reduction in the Interactive (Pre-Embedded) application type was due to the zero-skipping technique. This was ~21% in the case of the pre-embedded WikiMovies tasks. On the other hand, the $A^2$P-MANN inference reached, on average, 45.6% (44.8%) and 44.2% (42.5%) reductions for all the bAbI tasks in the Pre-Embedded (Interactive) application type in the Global and PerTask scenarios compared to the baseline network, respectively. Also, for the WikiMovies tasks, on average, our $A^2$P-MANN inference led to 60% (59%) computation reduction in the pre-embedded Global (PerTask) scenario. When the zero-skipping technique was used in the $A^2$P-MANN inference process on the bAbI tasks, an overall of 68% (59%) and 66% (57.5%) computation reductions for Interactive (Pre-Embedded) applications, in the Global and PerTask scenarios, respectively, were achieved. For the WikiMovies tasks, $A^2$P-MANN and zero-skipping (together) led to an overall of 75% (74%) computation reductions in the Global (PerTask) scenario. Generally, our proposed idea is more effective on interactive cases compared to the pre-embedded ones. This is due to the fact that, in the case of Interactive applications, the amount of computation required for the attention hops is the dominant part of the overall computation. Here, adding the input classifier network (ICN) leads to more computation reduction in the tasks that have a large number of easy inputs (most of the tasks) making

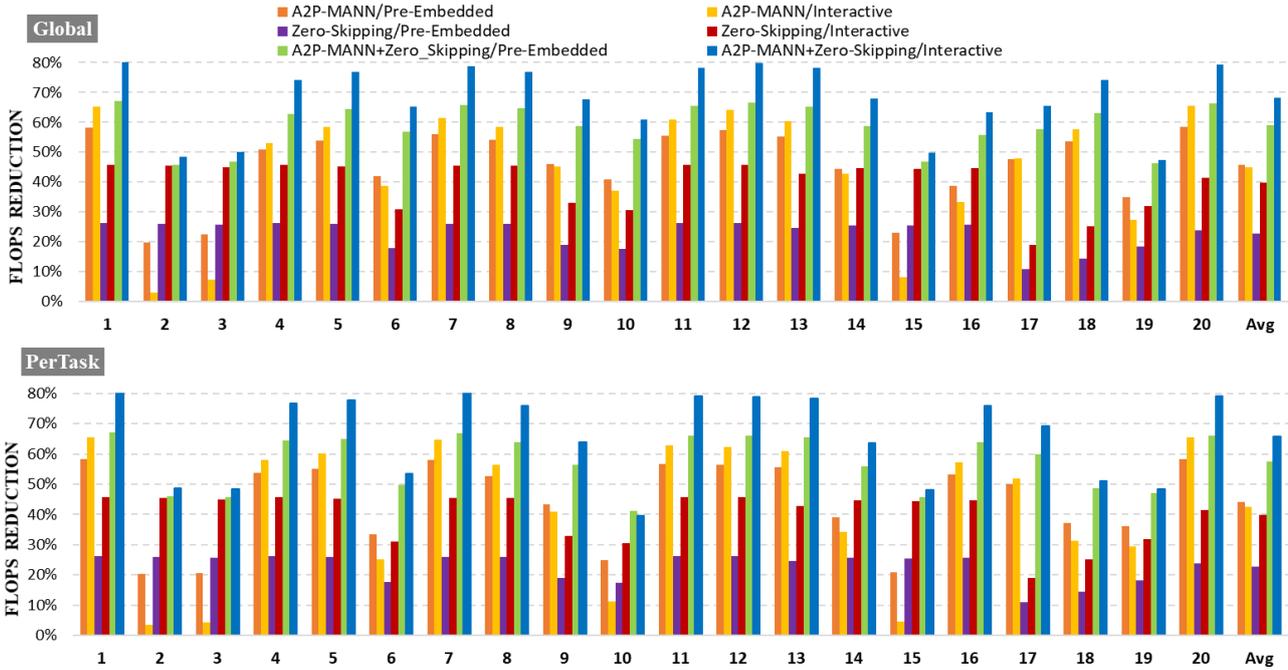

**Figure 9.** Computation reduction on 20 tasks of bAbI dataset using A²P-MANN inference (and zero-skipping) under Global and PerTask scenarios.

A²P-MANN more effective for most of the tasks in the interactive applications. For some tasks (*e.g.*, tasks 2 and 3 of bAbI dataset), however, a very small number of inputs are classified as easy (as indicated in Figure 6), and hence, ICN cannot help reducing the computations considerably. In those tasks, most of the computation reduction is achieved through pruning which is more efficient for pre-embedded applications. More specifically, in those application types, the FC layer computations form a higher part of the overall computations when compared to the case of the Interactive applications. In case of the Pre-Embedded WikiMovies tasks, because of the very large dictionary (*V*), a considerably higher amount of computations is required for the FC layer. This amount of computation forms a great portion of the overall computations. Our pruning method, therefore, is more effective for these tasks. Also, regarding the zero-skipping effectiveness in the case of interactive applications for bAbI tasks, we should mention that we avoided the re-embedding process for the skipped sentences (determined in the first hop) for the second hop as well as the following ones. In other words, after finding the sentences which have corresponding probability attention values less than a threshold ($\theta_{zs}$) in the first hop, we avoid embedding those sentences, again, in the remaining hops. Obviously, these sentences are redundant and should not have a considerable impact in the attention inference phase, for this particular question. This provides zero-skipping with more efficiency in the case of interactive applications. It should be mentioned, since the interactive applications was not considered in the evaluation, the avoidance of re-embedding was not addressed in the work where the zero-skipping technique was discussed ([17]).

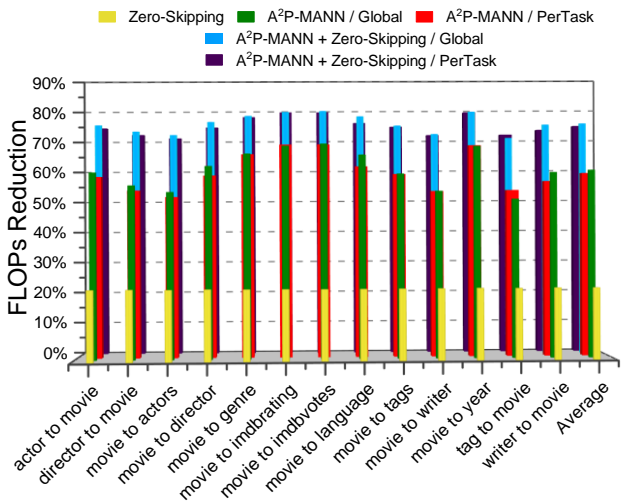

**Figure 10.** Computation reduction on WikiMovies tasks using A²P-MANN inference (and zero-skipping) under Global and PerTask scenarios.

***Platform Dependent Performance Evaluation:*** To compare the performance (*i.e.*, latency) of A²P-MANN to that of the baseline structures, we implemented them using PyTorch and executed them on the CPU (Intel Core i7-4790) and GPU (Nvidia 1080ti with 12GB memory). In this study, due to the similarity between the performance of the Global and PerTask scenarios, only the global confidence threshold was included. For the bAbI tasks, due to the small value of number of sentences in each story, $n_s$, (*i.e.*, 50) of the baseline network, the accurate execution time measurement was not possible. Thus, we have enlarged the $n_s$ of the considered network to 50K (5K) for the Pre-Embedded (Interactive) cases. A similar approach of enlarging the dataset by assuming an $n_s$ of 100M for showing the capability of the zero-skipping algorithm in reducing the runtime of MANN was used in [17]. Also, the zero-skipping technique was not used in the platform dependent study because of its ineffectiveness (and also even harmful) in some cases (especially, on GPUs) [17].

The normalized latencies of A²P-MANN with respect to the conventional MANN and KV-MANN on CPU and GPU platforms are illustrated in Figure 11 and Figure 12, respectively. As the figures indicate, on CPU, on average, a 40% (43%) latency reduction compared to the conventional MANN/bAbI for the Pre-Embedded (Interactive) application types for the A[2]P-MANN inference is achieved. On GPU, our proposed inference technique led to a 33% (41%) latency reduction, on average, compared to that of the conventional MANN/bAbI for Pre-Embedded (Interactive). Due to the large considered $n_s$, most of the overall computations belonged to the attention inference hops, and thus, most of the achieved latency reductions were owing to the utilization of ICN while the effect of the pruning techniques was negligible. For the 1st and 20th tasks, which had the maximum number of *Easy* inputs, the highest latency reductions of ~63% on CPU and 52% and 62% on GPU for Pre-Embedded and Interactive applications, respectively, were obtained. On the other hand, because of very few numbers of *Easy* input queries in the 2nd and 3rd tasks, the overhead of the ICN was large lowering the gain of the low computations of *Easy* inputs. Furthermore, our A[2]P-MANN inference method achieved 46% and 48% latency reduction, on average, for the KV-MANN/WikiMovies benchmark on the CPU and GPU, respectively. The high efficiency of our method on this benchmark originated from the fact that both the attention hops and the FC layer require significant computations providing the opportunity for the proposed methods (ICN and pruning) to significantly reduce the overall latency.

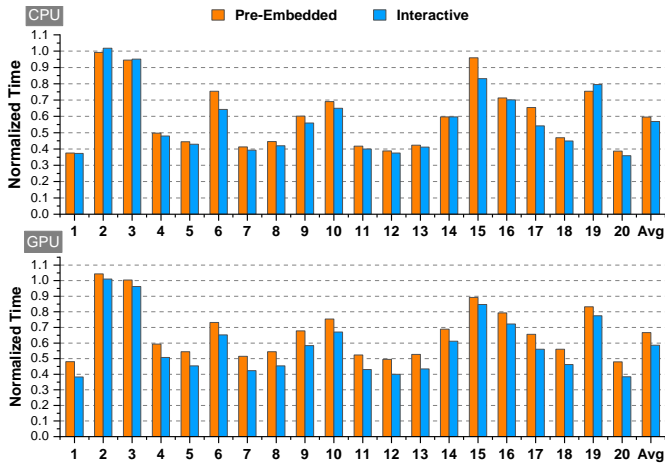

Figure 11. The normalized latency of A[2]P-MANN inference with respect to the conventional MANN on the CPU and GPU platforms.

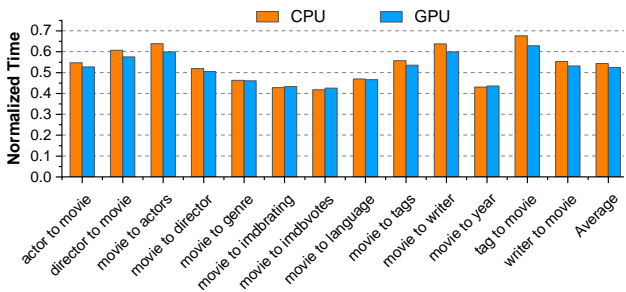

Figure 12. The normalized latency of A[2]P-MANN inference with respect to the KV-MANN/WikiMovies on the CPU and GPU platforms.

Finally, it should be mentioned that we used a smaller number of sentences for the Interactive applications of bAbI in the study owing to the fact that the database should be provided by the user in this case (*e.g.*, the contents of a book that the user has read, as stated in [17]). Obviously, embedding more sentences needs more time and energy. Thus, this limits the usage of the interactive approach for large databases. Nevertheless, to show the effectiveness of the proposed inference approach for large interactive applications, we repeated the study for $n_s = 50K$. The obtained timing results show, on average, about 44% latency reduction when using A[2]P-MANN. The reduction is larger than the values reported for the smaller databases.

## VI. CONCLUSION

In this paper, we proposed A[2]P-MANN which was a computationally efficient inference method. It dynamically controlled the number of required hops in memory-augmented neural networks. In its structure, a small additional input classifier network after the first hop determined whether the input is easy or hard. For easy inputs, the answer was extracted after the first hop using a retrained FC layer. For hard inputs, the calculations of remaining hops before using the final FC layer were performed. Also, to further reduce the computation in A[2]P-MANN, two methods for pruning the weight matrices of the FC layers were proposed. In the first one, *unused* rows of the final weight matrices found by analyzing the training labels were eliminated during inference. The second pruning method found *unimportant* rows by counting the number of small weights in each row. The suggested computation reduction techniques were independent from the hardware platform. The proposed approach was particularly suitable for resource-constrained environments such as embedded systems. The results obtained in this work showed about 42% computation reduction at the cost of less than 1% accuracy loss, on average, for A[2]P-MANN when compared to the conventional MANN for the bAbI dataset. Also, combined with zero-skipping, A[2]P-MANN achieved more than 57% (up to 68%) of computation reduction for this benchmark. Moreover, compared to the baseline KV-MANN structure, our methods reached 60% computation reduction, on average, for the WikiMovies dataset. Using zero-skipping in our proposed A[2]P-MANN, we reached a computation reduction of 75%. On the other hand, for the MANN/bAbI benchmark, A[2]P-MANN resulted in more than 40% latency reduction on a CPU while an average latency reduction of more than 33% (41% for interactive use cases) was achieved on a GPU. Finally, for the WikiMovies dataset, we achieved even higher latency reduction of 46% and 48% on the CPU and GPU platforms, respectively.

## REFERENCES

[1]     V. Sze, Y. H. Chen, T. J. Yang, and J. S. Emer, "Efficient Processing of Deep Neural Networks: A Tutorial and Survey," *Proceedings of the IEEE*, vol. 105, no. 12. Institute of Electrical and Electronics Engineers Inc., pp. 2295–2329, Dec. 01, 2017, doi: 10.1109/JPROC.2017.2761740.

[2]     S. Han, H. Mao, and W. J. Dally, "Deep Compression: Compressing Deep Neural Networks with Pruning, Trained Quantization and Huffman Coding," *4th Int. Conf. Learn. Represent. ICLR 2016 - Conf. Track Proc.*, Oct. 2015, Accessed: Jan. 16, 2021. [Online]. Available: https://arxiv.org/abs/1510.00149v5.

[3]     T.-J. Yang, Y.-H. Chen, and V. Sze, "Designing energy-efficient convolutional neural networks using energy-aware pruning," in *Proceedings of the IEEE Conference on Computer Vision and Pattern Recognition*, 2017, pp. 5687–5695.

[4]     S. Han, J. Pool, J. Tran, and W. J. Dally, "Learning both Weights

[5] E. Park *et al.*, "Big/little deep neural network for ultra low power inference," in *2015 International Conference on Hardware/Software Codesign and System Synthesis (CODES+ISSS)*, Oct. 2015, pp. 124–132, doi: 10.1109/CODESISSS.2015.7331375.

[6] H. Tann, S. Hashemi, R. I. Bahar, and S. Reda, "Runtime configurable deep neural networks for energy-accuracy trade-off," in *2016 International Conference on Hardware/Software Codesign and System Synthesis (CODES+ISSS)*, Oct. 2016, pp. 1–10.

[7] S. Venkataramani, A. Ranjan, K. Roy, and A. Raghunathan, "AxNN: Energy-efficient neuromorphic systems using approximate computing," in *2014 IEEE/ACM International Symposium on Low Power Electronics and Design (ISLPED)*, Aug. 2014, pp. 27–32, doi: 10.1145/2627369.2627613.

[8] S. Sukhbaatar, A. Szlam, J. Weston, and R. Fergus, "End-To-End Memory Networks," 2015.

[9] J. Weston, S. Chopra, and A. Bordes, "Memory Networks," *3rd Int. Conf. Learn. Represent. ICLR 2015 - Conf. Track Proc.*, Oct. 2014, Accessed: Jan. 16, 2021. [Online]. Available: http://arxiv.org/abs/1410.3916.

[10] A. Graves, G. Wayne, and I. Danihelka, "Neural Turing Machines," Oct. 2014, Accessed: Jan. 16, 2021. [Online]. Available: http://arxiv.org/abs/1410.5401.

[11] A. Graves *et al.*, "Hybrid computing using a neural network with dynamic external memory," *Nature*, vol. 538, no. 7626, pp. 471–476, Oct. 2016, doi: 10.1038/nature20101.

[12] Y. LeCun, "1.1 Deep Learning Hardware: Past, Present, and Future," in *2019 IEEE International Solid- State Circuits Conference - (ISSCC)*, Feb. 2019, vol. 2019-Febru, pp. 12–19, doi: 10.1109/ISSCC.2019.8662396.

[13] A. H. Miller, A. Fisch, J. Dodge, A. H. Karimi, A. Bordes, and J. Weston, "Key-value memory networks for directly reading documents," *EMNLP 2016 - Conf. Empir. Methods Nat. Lang. Process. Proc.*, pp. 1400–1409, 2016, doi: 10.18653/v1/d16-1147.

[14] J. Dodge *et al.*, "Evaluating Prerequisite Qualities for Learning End-to-End Dialog Systems," *4th Int. Conf. Learn. Represent. ICLR 2016 - Conf. Track Proc.*, Nov. 2015, Accessed: Jan. 16, 2021. [Online]. Available: http://arxiv.org/abs/1511.06931.

[15] F. Hill, A. Bordes, S. Chopra, and J. Weston, "The Goldilocks principle: Reading children's books with explicit memory representations," Nov. 2016, Accessed: Jan. 16, 2021. [Online]. Available: http://arxiv.org/abs/1511.02301.

[16] J. Weston, "Dialog-based Language Learning," *Adv. Neural Inf. Process. Syst.*, no. Nips, pp. 829–837, Apr. 2016, [Online]. Available: http://arxiv.org/abs/1604.06045.

[17] H. Jang, J. Kim, J.-E. Jo, J. Lee, and J. Kim, "MnnFast," in *Proceedings of the 46th International Symposium on Computer Architecture*, Jun. 2019, pp. 250–263, doi: 10.1145/3307650.3322214.

[18] S. Park, J. Jang, S. Kim, B. Na, and S. Yoon, "Memory-Augmented Neural Networks on FPGA for Real-Time and Energy-Efficient Question Answering," *IEEE Trans. Very Large Scale Integr. Syst.*, vol. 29, no. 1, pp. 162–175, 2021, doi: 10.1109/TVLSI.2020.3037166.

[19] V. V. Williams, "Breaking the Coppersmith-Winograd barrier," *Tensor*, pp. 1–71, 2011.

[20] J. R. Stevens, A. Ranjan, D. Das, B. Kaul, and A. Raghunathan, "Manna," in *Proceedings of the 52nd Annual IEEE/ACM International Symposium on Microarchitecture*, Oct. 2019, pp. 794–806, doi: 10.1145/3352460.3358304.

[21] S. Park, J. Jang, S. Kim, and S. Yoon, "Energy-Efficient Inference Accelerator for Memory-Augmented Neural Networks on an FPGA," in *2019 Design, Automation & Test in Europe Conference & Exhibition (DATE)*, Mar. 2019, pp. 1587–1590, doi: 10.23919/DATE.2019.8715013.

[22] J. Weston *et al.*, "Towards AI-Complete Question Answering: A Set of Prerequisite Toy Tasks," *4th Int. Conf. Learn. Represent. ICLR 2016 - Conf. Track Proc.*, Feb. 2015, Accessed: Jan. 16, 2021. [Online]. Available: http://arxiv.org/abs/1502.05698.

[23] O. Temam, "A defect-tolerant accelerator for emerging high-performance applications," in *2012 39th Annual International Symposium on Computer Architecture (ISCA)*, Jun. 2012, pp. 356–367, doi: 10.1109/ISCA.2012.6237031.

[24] C. Torres-Huitzil and B. Girau, "Fault and Error Tolerance in Neural Networks: A Review," *IEEE Access*, vol. 5, pp. 17322–17341, 2017, doi: 10.1109/ACCESS.2017.2742698.

[25] T.-W. Chin, C. Zhang, and D. Marculescu, "Layer-compensated Pruning for Resource-constrained Convolutional Neural Networks," *arXiv*, Sep. 2018, Accessed: Jan. 16, 2021. [Online]. Available: http://arxiv.org/abs/1810.00518.

[26] G. Gobieski, B. Lucia, and N. Beckmann, "Intelligence Beyond the Edge," in *Proceedings of the Twenty-Fourth International Conference on Architectural Support for Programming Languages and Operating Systems*, Apr. 2019, pp. 199–213, doi: 10.1145/3297858.3304011.

[27] B. Reagen *et al.*, "Ares: A framework for quantifying the resilience of deep neural networks," in *2018 55th ACM/ESDA/IEEE Design Automation Conference (DAC)*, Jun. 2018, pp. 1–6, doi: 10.1109/DAC.2018.8465834.

[28] M. Ali, A. Agrawal, and K. Roy, "RAMANN," in *Proceedings of the ACM/IEEE International Symposium on Low Power Electronics and Design*, Aug. 2020, pp. 61–66, doi: 10.1145/3370748.3406574.

[29] G. Huang, D. Chen, T. Li, F. Wu, L. van der Maaten, and K. Q. Weinberger, "Multi-Scale Dense Networks for Resource Efficient Image Classification," *arXiv*, Mar. 2017, Accessed: Jan. 16, 2021. [Online]. Available: http://arxiv.org/abs/1703.09844.

[30] P. Panda, A. Sengupta, and K. Roy, "Energy-Efficient and Improved Image Recognition with Conditional Deep Learning," *ACM J. Emerg. Technol. Comput. Syst.*, vol. 13, no. 3, pp. 1–21, May 2017, doi: 10.1145/3007192.

[31] O. Press and L. Wolf, "Using the Output Embedding to Improve Language Models," *15th Conf. Eur. Chapter Assoc. Comput. Linguist. EACL 2017 - Proc. Conf.*, vol. 2, pp. 157–163, Aug. 2016, Accessed: Jan. 16, 2021. [Online]. Available: http://arxiv.org/abs/1608.05859.

[32] J. Perez and F. Liu, "Gated End-to-End Memory Networks," *15th Conf. Eur. Chapter Assoc. Comput. Linguist. EACL 2017 - Proc. Conf.*, vol. 1, pp. 1–10, Oct. 2016, Accessed: Jan. 16, 2021. [Online]. Available: http://arxiv.org/abs/1610.04211.

[33] H. A. Thant, Khaing Moe San, Khin Mar Lar Tun, T. T. Naing, and N. Thein, "Mobile Agents Based Load Balancing Method for Parallel Applications," in *6th Asia-Pacific Symposium on Information and Telecommunication Technologies*, Nov. 2005, pp. 77–82, doi: 10.1109/APSITT.2005.203634.

[34] https://github.com/domluna/memn2n.

[35] https://github.com/facebook/MemNN/tree/master/KVmemnn.

[36] A. Paszke *et al.*, "PyTorch: An Imperative Style, High-Performance Deep Learning Library," in *Advances in Neural Information Processing Systems*, 2019, vol. 32, pp. 8026–8037, [Online]. Available:https://proceedings.neurips.cc/paper/2019/file/bdbca288fee7f92f2bfa9f7012727740-Paper.pdf.